\documentclass[conference,a4paper]{IEEEtran}

\usepackage{cite}
\usepackage[pdftex]{graphicx}
\usepackage{amsmath}
\usepackage{amssymb}
\usepackage{array}
\usepackage[caption=false,font=footnotesize]{subfig}
\begingroup\expandafter\expandafter\expandafter\endgroup
\expandafter\ifx\csname IncludeInRelease\endcsname\relax
  \usepackage{fixltx2e}
\fi
\usepackage{url}
\usepackage{booktabs}
\usepackage{array}
\usepackage{setspace}
\usepackage{bm} % bold math

\newcommand{\vect}[1]{\mathbf{#1}} % Bold letter
\DeclareMathSymbol{\R}{\mathalpha}{AMSb}{"52} % Real numbers
\begin{document}
\title{Improving Graph Convolutional Networks with Non-Parametric Activation Functions}

%\author{\IEEEauthorblockN{Simone Scardapane}
%\IEEEauthorblockA{DIET Dept.,\\Sapienza University, Italy\\
%simone.scardapane@uniroma1.it}
%\and
%\IEEEauthorblockN{Steven Van Vaerenbergh}
%\IEEEauthorblockA{Dept. Communications Engineering\\
%University of Cantabria, Spain\\
%steven.vanvaerenbergh@unican.es}
%\and
%\IEEEauthorblockN{Danilo Comminiello and Aurelio Uncini}
%\IEEEauthorblockA{DIET Dept.,\\Sapienza University, Italy\\
%danilo.comminiello@uniroma1.it}}

\author{\IEEEauthorblockN{Simone Scardapane\IEEEauthorrefmark{1},
Steven Van Vaerenbergh\IEEEauthorrefmark{2},
Danilo Comminiello\IEEEauthorrefmark{1} and
Aurelio Uncini\IEEEauthorrefmark{1}}
\IEEEauthorblockA{\IEEEauthorrefmark{1}\textit{Dept. Information Eng., Electronics and Telecomm., Sapienza University of Rome, Italy.}} %
\IEEEauthorblockA{\IEEEauthorrefmark{2}\textit{Dept. Communications Eng., University of Cantabria, Spain.}}
Corresponding author email: simone.scardapane@uniroma1.it}


\maketitle

\begin{abstract}
Graph neural networks (GNNs) are a class of neural networks that allow to efficiently perform inference on data that is associated to a graph structure, such as, e.g., citation networks or knowledge graphs. While several variants of GNNs have been proposed, they only consider simple nonlinear activation functions in their layers, such as rectifiers or squashing functions. In this paper, we investigate the use of graph convolutional networks (GCNs) when combined with more complex activation functions, able to adapt from the training data. More specifically, we extend the recently proposed kernel activation function, a non-parametric model which can be implemented easily, can be regularized with standard $\ell_p$-norms techniques, and is smooth over its entire domain. Our experimental evaluation shows that the proposed architecture can significantly improve over its baseline, while similar improvements cannot be obtained by simply increasing the depth or size of the original GCN.
\end{abstract}


\IEEEpeerreviewmaketitle

\section{Introduction}
\label{sec:intro}

Efficient processing of graph-structured data (e.g., citation networks) has a range of different applications, going from bioinformatics to text analysis and sensor networks, among others. Of particular importance is the design of learning methods that are able to take into account both numerical characteristics of each node in the graph and their inter-connections \cite{sandryhaila2014big}. While graph machine learning techniques have a long history, recently we witness a renewed interest in models that are defined and operate directly in the graph domain (as compared to a Euclidean space), instead of including the graph information only a posteriori in the optimization process (e.g., through manifold regularization techniques \cite{yuan2015scene}), or in a pre-processing phase via graph embeddings \cite{yang2016revisiting}. Examples of native graph models include graph linear filters \cite{di2016adaptive}, their kernel counterparts \cite{romero2017kernel}, and graph neural networks (GNNs) \cite{bronstein2017geometric}.

GNNs are particularly interesting because they promise to bring the performance of deep learning models \cite{goodfellow2016deep} to graph-based domains. In particular, convolutional neural networks (CNNs) are nowadays the de-facto standard for processing image data. CNNs exploit the image structure by performing spatial convolutions (with a limited receptive field) on the image, thus increasing parameter sharing and lowering their complexity. A number of authors recently have explored the possibility of extending CNNs to the graph domain by several generalizations of the convolution operator, a trend which has been generically called `geometric deep learning' \cite{bronstein2017geometric}. In one of the first proposals \cite{bruna2014spectral}, graph Fourier transform was used at every layer of the network to perform filtering operations in a graph-frequency domain. However, this approach was not scalable as it scaled linearly with the size of the graph. Defferard et al. \cite{defferrard2016convolutional} extended this approach by using polynomial filters on the frequency components, which can be rewritten directly in the graph domain, avoiding the use of the graph Fourier transform. A further set of modifications was proposed by Kipf and Welling \cite{kipf2017semi} (described more in depth in Section \ref{sec:graph_convolutional_networks}), resulting in a generic graph convolutional network (GCN). GCN has been successfully applied to several real-world scenarios, including semi-supervised learning \cite{kipf2017semi}, matrix completion \cite{berg2017graph}, program induction \cite{allamanis2017learning}, modeling relational data \cite{schlichtkrull2017modeling}, and several others.

Like most neural networks, GCNs interleave linear layers, wherein information is adaptively combined according to the topology of the graph, with nonlinear activation functions that are applied element-wise. The information over the nodes can then be combined to obtain a graph-level prediction, or kept separate to obtain a node-specific inference (e.g., predictions on unlabeled nodes from a small set of labeled ones). Most literature for GCNs has worked with very simple choices for the activation functions, such as the rectified linear unit (ReLU) \cite{kipf2017semi}. However, it is well known that the choice of the function can have a large impact on the final performance of the neural network. Particularly, there is a large literature for standard neural networks on the design of flexible schemes for \textit{adapting} the activation functions themselves from the training data \cite{he2015delving,goodfellow2013maxout,agostinelli2014learning,scardapane2017kafnets}. These techniques range from the use of simple parametrizations over known functions (e.g., the parametric ReLU \cite{he2015delving}), to the use of more sophisticated non-parametric families, including the Maxout network \cite{goodfellow2013maxout}, the adaptive piecewise linear unit \cite{agostinelli2014learning}, and the recently proposed kernel activation function (KAF) \cite{scardapane2017kafnets}.

\subsubsection*{Contribution of the paper}

In this paper, we conjecture that choosing properly the activation function (beyond simple ReLUs) can further improve the performance of GCNs, possibly by a significant margin. To this end, we enhance the basic GCN model by extending KAFs for the activation functions of the filters. Each KAF models a one-dimensional activation function in terms of a simple kernel expansion (see the description in Section \ref{sec:proposed_gcn}). By properly choosing the elements of this expansion beforehand, we can represent each function with a small set of (linear) mixing coefficients, which are adapted together with the weights of the convolutional layers using standard back-propagation. Apart from flexibility, the resulting scheme has a number of advantages, including smoothness in its domain and possibility of implementing the algorithm using highly vectorized GPU operations. We compare on two benchmarks for semi-supervised learning, showing that the proposed GCN with KAFs can significantly outperform all competing baselines with a marginal increase in computational complexity (which is offset by a faster convergence in terms of epochs).

\subsubsection*{Structure of the paper}

Section \ref{sec:graph_convolutional_networks} introduces the problem of inference over graphs and the generic GCN model. The proposed extension with KAFs is described in Section \ref{sec:proposed_gcn}. We evaluate and compare the model in Section \ref{sec:experiments} before concluding in Section \ref{sec:conclusions}.

\section{Graph convolutional networks}
\label{sec:graph_convolutional_networks}

\subsection{Problem setup}

Consider a generic undirected graph $\mathcal{G} = \left(\mathcal{V}, \mathcal{E}\right)$, where $\mathcal{V} = \left\{1, \ldots, N\right\}$ is the set of $N$ vertices, and $\mathcal{E} \subseteq \mathcal{V} \times \mathcal{V}$ is the set of edges connecting the vertices. The graph is equivalently described by the (weighted) adjacency matrix $\vect{A} \in \R^{N \times N}$, where its generic element $a_{ij} \geq 0$ if and only if nodes $i$ and $j$ are connected. A graph signal \cite{sandryhaila2014big} is a function $f: \mathcal{V} \rightarrow \R^F$ associating to vertex $n$ a $F$-dimensional vector $\vect{x}_n$ (each dimension being denoted as a channel). For a subset $\mathcal{L} \subset \mathcal{V}$ of the vertices we have also available a node-specific label $y_n$, which can be either a real number (graph regression) or a discrete quantity (graph classification). Our task is to find the correct labels for the (unlabeled) nodes that are not contained in $\mathcal{L}$. Note that if the graph is unknown and/or must be inferred from the data, this setup is equivalent to standard semi-supervised (or, more precisely, transductive) learning \cite{Chapelle:2010:SL:1841234}.

Ignoring for a moment the graph information, we could solve the problem by training a standard (feedforward) NN to predict the label $y$ from the input $\vect{x}$ \cite{goodfellow2016deep}. A NN is composed by stacking $L$ layers, such that the operation of the $l$th layer can be described by:
\begin{equation}
\vect{h}_l = g \left( \vect{W}_l\vect{h}_{l-1} \right) \,,
\label{eq:nn_layer}
\end{equation}
where $\vect{h}_{l-1} \in \mathbb{R}^{N_{l-1}}$ is the input to the layer, $\vect{W}_l \in \mathbb{R}^{N_{l-1} \times N_l}$ are trainable weights (ignoring for simplicity any bias term), and $g(\cdot)$ is a element-wise nonlinear function known as activation function (AF). The NN takes as input $\vect{x} = \vect{h}_0$, providing $\hat{\vect{y}} = \vect{h}_L \in \R^C$ as the final prediction. A number of techniques can be used to include unlabeled information in the training process of NNs, including ladder networks \cite{rasmus2015semi}, pseudo-labels \cite{lee2013pseudo}, manifold regularization \cite{yuan2015scene}, and neural graph machines \cite{bui2017neural}. As we stated in the introduction, however, the aim of GNNs is  to include the proximity information contained in $\vect{A}$ directly \textit{inside} the processing layers of NNs, in order to further improve the performance of such models. As shown in the experimental section, working directly in the graph domain can obtain vastly superior performances as compared to standard semi-supervised techniques.

Note that if the nodes of the graphs are organized in a regular grid, spatial information can be included by adding standard convolutional operations to \eqref{eq:nn_layer}, as is common in CNNs \cite{goodfellow2016deep}. In order to extend this idea to general unweighted graphs, we need some additional tools from the theory of graph signal processing.

\subsection{Graph Fourier transform and graph neural networks}

In order to define a convolutional operation in the graph domain, we can exploit the normalized Laplacian matrix $\vect{L} = \vect{I} - \vect{D}^{-\frac{1}{2}}\vect{A}\vect{D}^{-\frac{1}{2}}$, where $\vect{D}$ is a diagonal matrix with $D_{nn} = \sum_{t=1}^N A_{tt}$. We denote the eigendecomposition of $\vect{L}$ as $\vect{L}=\vect{U}\vect{\Lambda}\vect{U}^T$, where $\vect{U}$ is a matrix collecting column-wise the eigenvectors of $\vect{L}$, and $\vect{\Lambda}$ is a diagonal matrix with associated eigenvalues. We further denote by $\vect{X} \in \R^{N \times F}$ the matrix collecting the input signal across all nodes, and by $\vect{X}_i \in \R^N$ its $i$th column. The graph Fourier transform of $\vect{X}_i$ can be defined as \cite{sandryhaila2014big}:
\begin{equation}
\hat{\vect{X}}_i = \vect{U}^T\vect{X}_i \,,
\end{equation}
and, because the eigenvectors form an orthonormal basis, we can also define an inverse Fourier transform as $\vect{X}_i = \vect{U}\hat{\vect{X}}_i$. We can exploit the graph Fourier transform and define a  convolutional layer operating on graphs as \cite{bruna2014spectral}:
\begin{equation}
\vect{H}_{1} =  g \left( \sum_{i=1}^F \vect{U} h(\vect{\Lambda}; \vect{\Theta}_i) \vect{U}^T \vect{X}_i \right) \,,
\label{eq:generic_gcn_layer}
\end{equation}
where $h(\vect{\Lambda}; \vect{\Theta}_i)$ is a (channel-wise) filtering operation acting on the eigenvalues, having adaptable parameters $\vect{\Theta}_i$. The previous operation can then be iterated to get successive representations $\vect{H}_2, \vect{H}_3, \ldots$ up to the final node-specific predictions. The choice of $h(\cdot)$ determines the complexity of training. In particular, \cite{defferrard2016convolutional} proposed to make the problem tractable by working with polynomial filters over the eigenvalues:

\begin{equation}
h(\vect{\Lambda}; \vect{\Theta}_i) = \sum_{k=0}^{K-1} \Theta_{ik} T_k(\widetilde{\vect{\Lambda}}) \,,
\label{eq:polynomial_filter}
\end{equation}
where $\widetilde{\vect{\Lambda}} = 2\vect{\Lambda}/\lambda_{\text{max}} - \vect{I}$ (with $\lambda_{\text{max}}$ being the largest eigenvalue of $\vect{L}$), and $T_k(\cdot)$ the Chebyshev polynomial of order $k$ defined by the recurrence relation:
\begin{equation}
T_k(\widetilde{\vect{\Lambda}}) = 2\widetilde{\vect{\Lambda}}T_{k-1}(\widetilde{\vect{\Lambda}}) - T_{k-2}(\widetilde{\vect{\Lambda}}) \,,
\end{equation}
with $T_0(\widetilde{\vect{\Lambda}}) = 1$ and $T_1(\widetilde{\vect{\Lambda}}) = \widetilde{\vect{\Lambda}}$. The filtering operation defined by \eqref{eq:polynomial_filter} is advantageous because each filter is parameterized with only $K$ values (with $K$ chosen by the user). Additionally, it is easy to show that the filter is localized over the graph, in the sense that the output at a given node depends only on nodes up to maximum $K$ hops from it. Thanks to the choice of a polynomial filter, we can also avoid the expensive multiplications by $\vect{U}$ and $\vect{U}^T$ by rewriting the filtering operation directly on the original graph domain \cite{defferrard2016convolutional}.

In order to avoid the need for the recurrence relation, \cite{kipf2017semi} proposed to further simplify the expression by setting $K=1$, $\theta_{i0} = - \theta_{i1}$ for all channels, and assuming $\lambda_{\text{max}} = 2$. Back-substituting in the original expression \eqref{eq:generic_gcn_layer} we obtain:
\begin{equation}
\vect{H}_1 = g \left( \left( \vect{I} + \vect{D}^{-\frac{1}{2}}\vect{A}\vect{D}^{-\frac{1}{2}} \right) \vect{X} \vect{\Theta}_1 \right) \,,
\end{equation}
where $\vect{\Theta}_1 \in \R^{F \times N_1}$ is the matrix of adaptable filter coefficients. Practically, we can further avoid some numerical instabilities (due to the range of the eigenvalues) by substituting $\vect{I} + \vect{D}^{-\frac{1}{2}}\vect{A}\vect{D}^{-\frac{1}{2}}$ with $\widetilde{\vect{D}}^{-\frac{1}{2}}\widetilde{\vect{A}}\widetilde{\vect{D}}^{-\frac{1}{2}}$, where $\widetilde{\vect{A}} = \vect{A} + \vect{I}$ and $\widetilde{D}_{nn} = \sum_{t=1}^N \widetilde{A}_{tt}$. The previous expression can then be iterated as in \eqref{eq:nn_layer} to obtain a graph convolutional network (GCN). As an example, a GCN with two layers is given by:
\begin{equation}
\hat{\vect{Y}} = g \left( \vect{\widehat{A}} g \left( \vect{\widehat{A}}\vect{X}\vect{\Theta}_1 \right) \vect{\Theta}_2 \right) \,,
\label{eq:two_layer_gcn}
\end{equation}
where we defined $\vect{\widehat{A}} = \widetilde{\vect{D}}^{-\frac{1}{2}}\widetilde{\vect{A}}\widetilde{\vect{D}}^{-\frac{1}{2}}$. The parameters $\left\{\vect{\Theta}_l\right\}_{l=1}^L$ can be initialized randomly and trained by minimizing a suitable loss over the labeled examples, such as the cross-entropy for classification problems \cite{kipf2017semi}. A single layer of the GCN can combine information coming only from the immediate neighbors of a node, while using multiple layers allows information to flow over several hops. In practice, two layers are found to be sufficient for many benchmark problems, and further depth does not provide a benefit \cite{kipf2017semi}. Note, however, that the previous expression requires to choose a proper activation function. This is the subject of the next section.

\section{Proposed GCN with kernel activation functions}
\label{sec:proposed_gcn}

Up to this point, we assumed that the activation functions $g(\cdot)$ in \eqref{eq:two_layer_gcn} were given. Note that the functions for different layers (or even for different filters in the same layer) need not be the same. Particularly, the outer function is generally chosen in a task-dependent fashion (e.g., softmax for classification \cite{goodfellow2016deep}). However, choosing different activation functions for the hidden layers can vastly change the performance of the resulting models and their flexibility.

Most of the research on GNNs and GCNs has focused on simple activation functions, such as the ReLU function:

\begin{equation}
g(s) = \max\left(0, s\right) \,,
\end{equation}
where $s$ is a generic element on which the function is applied. It is easy to add a small amount of flexibility by introducing a simple parametrization of the function. For example, the parametric ReLU \cite{he2015delving} adds an adaptable slope $\alpha$ (independent for every filter) to the negative part of the function:

\begin{equation}
g(s) = \min\left(0, -\alpha s \right) + \max\left(0, s\right) \,.
\end{equation}
As we stated in the introduction, such parametric functions might not provide a significant increase in performance in general, and a lot of literature has been devoted to designing non-parametric activation functions that can adapt to a very large family of shapes. A simple technique would be to project the activation $s$ to a high-dimensional space $\phi(s)$, and then adapt a different linear function on this feature space for every filter. However, this method becomes easily unfeasible for a large dimensionality of $\phi(\cdot)$.

The idea of KAFs \cite{scardapane2017kafnets} is to avoid the expensive feature computation by working instead with a kernel expansion. A known problem of kernel methods is that the elements that we use for computing the kernel values (what is called the dictionary in the kernel filtering literature \cite{liu2011kernel}) can be extremely hard to select. The insight in \cite{scardapane2017kafnets} is to exploit the fact that we are working with one-dimensional functions, by fixing the dictionary beforehand by sampling uniformly $D$ elements (with $D$ chosen by the user) around zero, and only adapting the linear coefficients of the kernel expansion (see the equations below). In this way, the parameter $D$ controls the flexibility of our approach: by increasing it, we increase the flexibility of each filter at the cost of a larger number of parameters per-filter.

More formally, we propose to model each activation function inside the GCN as follows:

\begin{equation}
g(s) = \sum_{i=1}^D \alpha_i \kappa\left(s, d_i\right) \,,
\label{eq:kaf}
\end{equation}
where $d_i$ are the elements of the dictionary selected according to before, while the mixing coefficients $\left\{\alpha_i\right\}_{i=1}^D$ are initialized at every filter and adapted independently together with $\left\{\vect{\Theta}_l\right\}_{l=1}^L$. $\kappa(\cdot, \cdot)$ in \eqref{eq:kaf} is a generic kernel function, and we use the 1D Gaussian kernel in our experiments:
\begin{equation}
\kappa(s, d_i) = \exp\left\{-\gamma\left(s - d_i\right)^2\right\} \,,
\label{eq:gaussian_kernel}
\end{equation}
where $\gamma \in \R$ is a free parameter. Since in our context the effect of $\gamma$ only depends on the sampling of the dictionary, we select it according to the rule-of-thumb proposed in \cite{scardapane2017kafnets}:

\begin{equation}
\gamma = \frac{1}{6\Delta^2} \,,
\label{eq:gamma_rule_of_thumb}
\end{equation}
where $\Delta$ is the distance between two elements of the dictionary. 

\begin{figure}
\centering
\includegraphics[width=0.80\columnwidth,keepaspectratio]{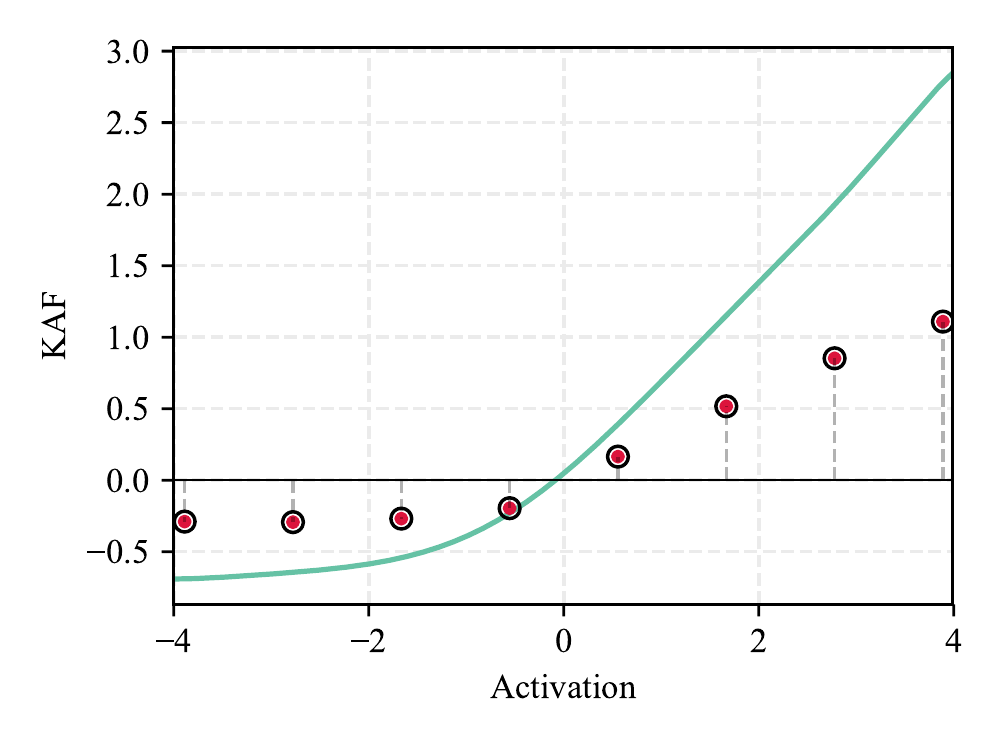}
\caption{Example of initializing a KAF according to the ELU function, shown with $D=8$ and $\gamma$ chosen according to \eqref{eq:gamma_rule_of_thumb}.}
\label{fig:kaf_initialization_elu}
\end{figure}

In order to avoid numerical problems during the initial phase of learning, we initialize the mixing coefficients to mimic as close as possible an exponential linear unit (ELU) following the strategy in \cite{scardapane2017kafnets}. Denoting by $\vect{t}$ a sampling of the ELU at the same positions as the dictionary, we initialize the coefficients as:
\begin{equation}
\boldsymbol{\alpha} = \left(\vect{K} + \varepsilon\vect{I}\right)^{-1}\vect{t} \,,
\label{eq:kaf_initialization_krr}
\end{equation}
where $\boldsymbol{\alpha}$ is the vector of mixing coefficients, $\vect{K} \in \R^{D \times D}$ is the kernel matrix computed between $\vect{t}$ and the dictionary, and we add a diagonal term with $\varepsilon > 0$ to avoid degenerate solutions with very large mixing coefficients (see Fig. \ref{fig:kaf_initialization_elu} for an example).

For more details on \eqref{eq:kaf} we refer to the original publication \cite{scardapane2017kafnets}. Here, we briefly comment on some advantages of using this non-parametric formulation. First of all, \eqref{eq:kaf} can be implemented easily in most deep learning libraries using vectorized operations (more on this later on in the experimental section). Second, the functions defined by \eqref{eq:kaf} are smooth over their entire domain. Finally, the mixing parameters $\left\{\alpha_i\right\}_{i=1}^D$ can be handled like all other parameters, particularly by applying standard regularization techniques (e.g., $\ell_2$ or $\ell_1$ regularization).

\section{Experimental results}
\label{sec:experiments}

\subsection{Experimental setup}

We evaluate the proposed model on two semi-supervised learning benchmarks taken from \cite{kipf2017semi}, whose characteristics are reported in Table \ref{tab:datasets}. Both represents citation networks, where vertices are documents (each one represented by a bag-of-words), edges are links, and each document can belong to a given class. The last column in Table \ref{tab:datasets} is the percentage of labeled nodes in the datasets.

The baseline network is a two-layer GCN as in \eqref{eq:two_layer_gcn}, while in our proposed method we simply replace the inner activation functions with KAFs, with a dictionary of $D=20$ elements sampled uniformly in $\left[-2.0, 2.0\right]$. We optimize a cross-entropy loss given by:
\begin{equation}
L = \sum_{l \in \mathcal{L}} \sum_{c=1}^C Y_{lc} \log\left(\widehat{Y}_{lc}\right) \,.
\end{equation}
The loss is optimized using the Adam algorithm \cite{goodfellow2016deep} in a batch fashion. The labeled dataset is split following the same procedure as \cite{kipf2017semi}, where we use $20$ nodes per class for training, $1000$ elements for validation, and the remaining elements for testing the accuracy. The validation set is used for evaluating the loss at each epoch, and we stop as soon as the loss is not decreasing with respect to the last $10$ epochs. All hyper-parameters are selected in accordance with \cite{kipf2017semi} as they were already optimized for the Cora dataset: $16$ filters in the hidden layer, an initial learning rate of $0.01$, and an additional $\ell_2$ regularization on the weights with weighting factor $5e^{-4}$. In addition, we use dropout (i.e., we randomly remove filters from the hidden layer during training) with probability $0.5$. The splits for the datasets are taken from \cite{yang2016revisiting}, and we average over $15$ different initializations for the networks. 

For the implementation, we extend the original code for the GCN.\footnote{\url{https://github.com/tkipf/pygcn}} Specifically, multiplication with the adjacency matrix in \eqref{eq:two_layer_gcn} is done using efficient sparse data structures, allowing to perform it in linear time with respect to the number of edges. All experiments are run in Python using a Tesla K80 as backend.

\begin{table}
\caption{Brief description of the datasets (for more details see the text and \cite{yang2016revisiting,kipf2017semi}).}
{\centering\hfill{}
	\setlength{\tabcolsep}{4pt}
	\renewcommand{\arraystretch}{1}
	\begin{tabular}{lccccc}   %@{}p{0.3\columnwidth}p{0.2\columnwidth}p{0.2\columnwidth}@{}
	\toprule
	\textbf{Dataset} & \textbf{Nodes} & \textbf{Edges} & \textbf{Classes} & \textbf{Features} & \textbf{Labels}\\ 
	\midrule
	Citeseer & $3327$ & $4732$ & $6$ & $3703$ & $3.60 \%$ \\
	Cora & $2708$ & $5429$ & $7$ & $1433$ & $5.20 \%$ \\
	\bottomrule
	\end{tabular}
}
\hfill{}
\label{tab:datasets}
\end{table}

\subsection{Experimental results}

\begin{table*}
\caption{Results in terms of accuracy over the test set. All baselines are taken from \cite{yang2016revisiting,kipf2017semi}. The proposed algorithm is denoted as KAF-GCN. Best results are highlighted in bold.}
{\centering\hfill{}
	\setlength{\tabcolsep}{4pt}
	\renewcommand{\arraystretch}{1}
	\begin{tabular}{lcccccccc}   %@{}p{0.3\columnwidth}p{0.2\columnwidth}p{0.2\columnwidth}@{}
	\toprule
	\textbf{Dataset} & \textbf{ManiReg} & \textbf{SemiEmb} & \textbf{LP} & \textbf{DeepWalk} & \textbf{ICA} & \textbf{Planetoid} & \textbf{GCN} & \textbf{KAF-GCN}\\ 
	\midrule
	Citeseer & $60.1$ & $59.6$ & $45.3$ & $43.2$ & $69.1$ & $67.7$ & $70.3$ & $\vect{70.9 \pm 0.1}$ \\
	Cora & $59.5$ & $29.0$ & $68.0$ & $67.2$ & $75.1$ & $75.7$ & $81.5$ & $\vect{83.0 \pm 0.2}$ \\
	\bottomrule
	\end{tabular}
}
\hfill{}
\label{tab:results}
\end{table*}

The results (in terms of accuracy computed on the test set) of GCN and the proposed GCN with KAFs (denoted as KAF-GCN) are given in Table \ref{tab:results}. We also report the performance of six additional baseline semi-supervised algorithms taken from \cite{yang2016revisiting,kipf2017semi}, including manifold regularization (ManiReg), semi-supervised embedding (SemiEmb), label propagation (LP), skip-gram based graph embeddings (DeepWalk), iterative classification (ICA), and Planetoid \cite{kipf2017semi}. References and a full description of all the baselines can be found in the original papers.

Note how, in both cases, the proposed KAF-GCN is able to outperform all the other baselines in a stable fashion (for KAF-GCN, we also report the standard deviation with respect to the different weights' initialization). In particular, results on the Cora dataset represent (to the best of our knowledge) the state-of-the-art results when using such a small labeled dataset. It is important to underline that the gap in performance between GCN and KAF-GCN cannot be reduced by merely increasing the depth (or size) of the former, since its architecture is already fine-tuned to the specific datasets. In particular, \cite[Appendix B]{kipf2017semi} reports results showing that neither GCN, nor a variant with residual connections can obtain a higher accuracy when adding more layers. In our opinion, this points to the importance of having adaptable activation functions able to more efficiently process the information coming from the different filters. Due to a lack of space, we are not able to show the shapes of the functions resulting from the optimization process, although they are found to be similar to those obtained by \cite{scardapane2017kafnets}, and the interested reader is referred there.

\begin{figure}
\centering
\includegraphics[width=0.90\columnwidth,keepaspectratio]{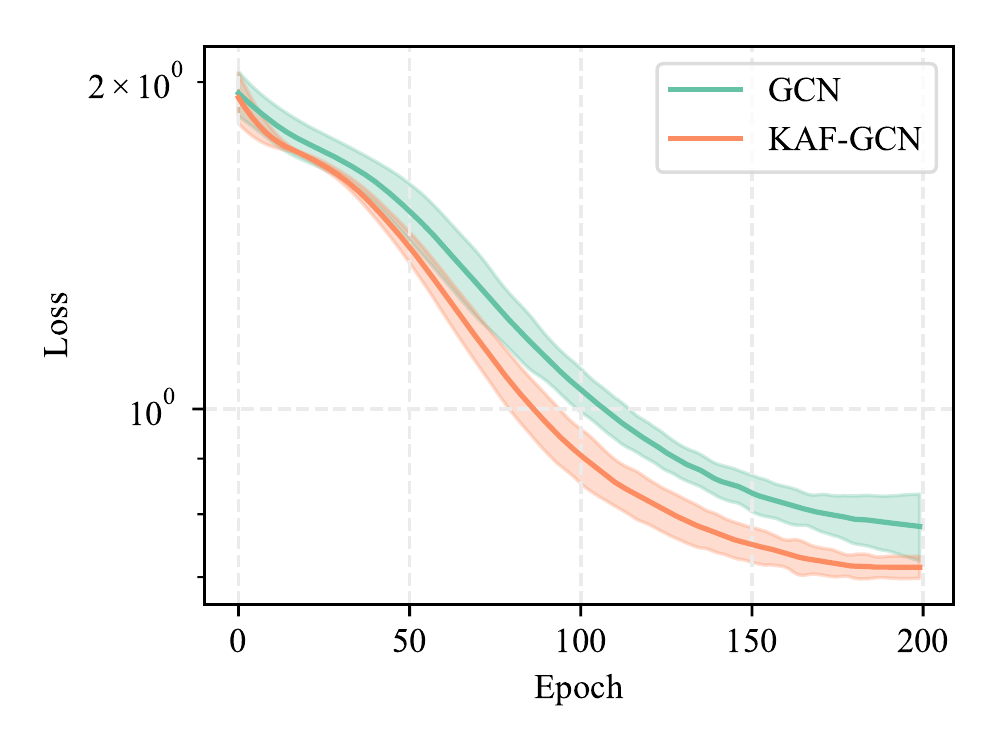}
\caption{Loss evolution for GCN and KAF-GCN on the Cora dataset. We show the mean with a solid line and the standard deviation with a lighter color.}
\label{fig:convergence}
\end{figure}

In terms of training time, our implementation of KAF-GCN requires roughly 10\% more computation per epoch than the standard GCN for the Cora dataset, and 15\% more for the Citeseer one. However, we show in Fig. \ref{fig:convergence} the (average) loss evolution on the Cora dataset for the two models. We see that KAF-GCN generally converges faster than GCN, possibly due to the higher flexibility allowed by the network. Due to this, KAF-GCN requires a lower number of epochs to achieve convergence (as measured by the early stopping criterion), requiring on average $25/30$ epochs less to converge, which more than compensate for the increased computational time per-epoch.

\section{Conclusions}
\label{sec:conclusions}

In this paper we have shown that the performance of graph convolutional networks can be significantly improved with the use of more flexible activation functions for the hidden layers. Specifically, we evaluated the use of kernel activation functions on two semi-supervised benchmark tasks, resulting in faster convergence and higher classification accuracy.

Two limitations of the current work are the use of undirected graph topologies, and the training in a batch regime. While \cite{kipf2017semi} handles the former case by rewriting a directed graph as an equivalent (undirected) bipartite graph, this operation is computationally expensive. We plan on investigating recent developments on graph Fourier transforms for directed graphs \cite{sardellitti2017graph} to define a simpler formulation. More in general, we aim to test non-parametric activation functions for other classes of graph neural networks (such as gated graph neural networks), which would allow to process sequences of graphs or multiple graphs at the same time.

\section*{Acknowledgment}

Simone Scardapane was supported in part by Italian MIUR, GAUChO project, under Grant 2015YPXH4W\_004.



\bibliographystyle{IEEEtran}
\bibliography{biblio}

\end{document}